\documentclass[10pt,twocolumn,letterpaper]{article}

\usepackage{cvpr}
\usepackage{times}
\usepackage{epsfig}
\usepackage{graphicx}
\usepackage{amsmath}
\usepackage{amssymb}
\usepackage{caption}
\usepackage{multirow}
\usepackage{booktabs}
\usepackage{subcaption}
\usepackage[font=small,labelfont=bf]{caption}

\usepackage[breaklinks=true,bookmarks=false]{hyperref}

\cvprfinalcopy 

\def\cvprPaperID{083} 

\begin{document}
\renewcommand*{\thefootnote}{\fnsymbol{footnote}}
\setcounter{footnote}{0}

\title{SuperTran: Reference Based Video Transformer \\
for Enhancing Low Bitrate Streams in Real Time}

\author{ Tejas Khot\footnotemark  \qquad
Nataliya Shapovalova\footnotemark  \qquad
Silviu Andrei\footnotemark[\value{footnote}]  \qquad
Walterio Mayol-Cuevas\\
Amazon
}


\twocolumn[{%
\maketitle

    \vspace{-1cm}
    \renewcommand\twocolumn[1][]{#1}%
    \renewcommand\tabcolsep{0.5pt}
	\begin{center}
		\centering
        {\def\arraystretch{0.5}
\begin{tabular}{ccccc}
    \includegraphics[width=\textwidth]{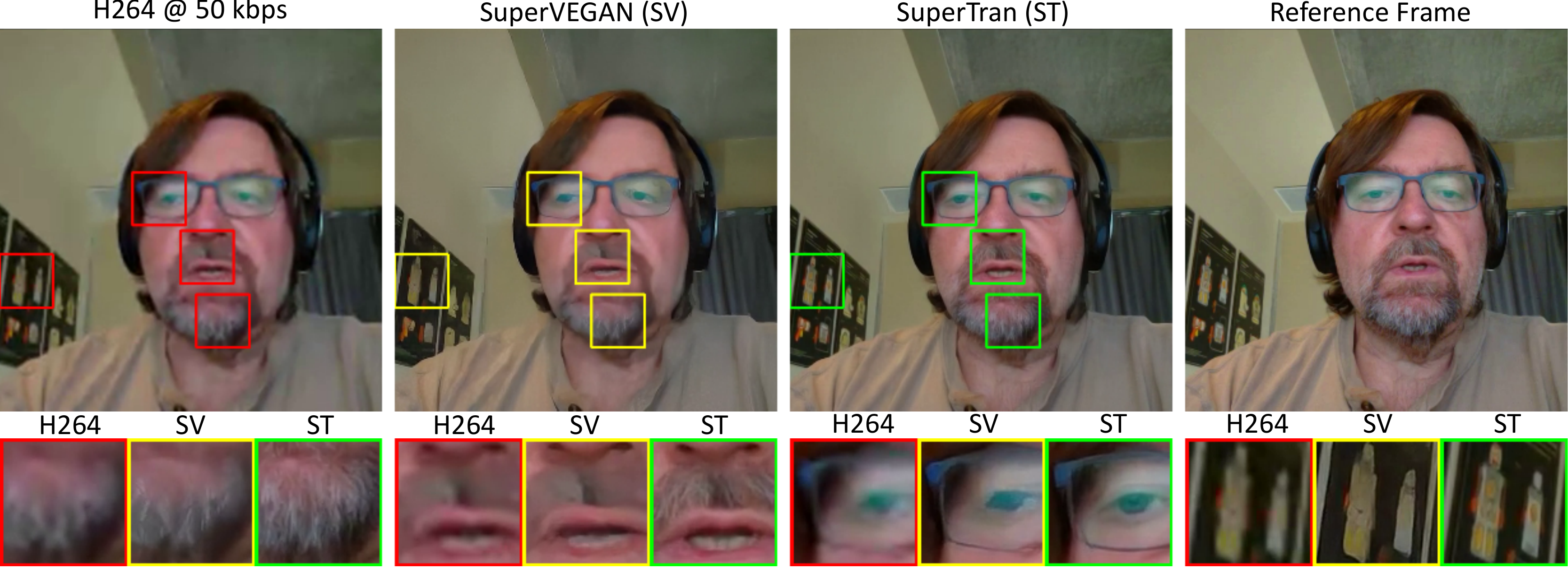}\\
    \includegraphics[width=\textwidth]{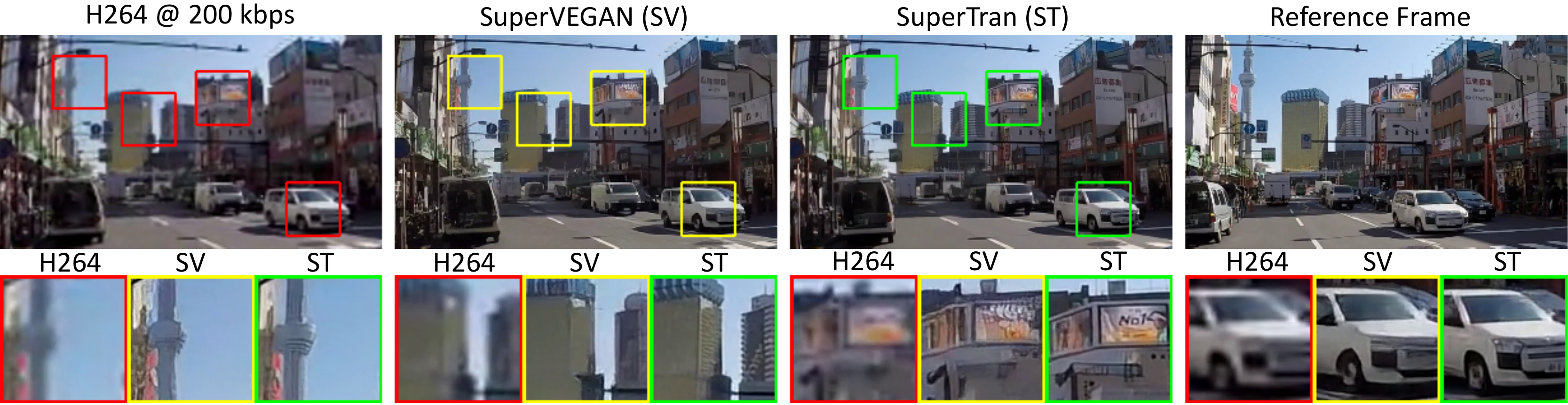}
    
\end{tabular}
} 
\vspace{-1em}
\captionof{figure}{
    \textbf{Qualitative result comparison.} 
    Comparison of results on two scenarios at low bitrates (50 and 200 kbps). 
    The H264 frame is fed as input to SuperVEGAN\cite{supervegan} and SuperTran. SuperTran also consumes the reference frame. 
    SuperTran is able to borrow details from the reference frame under change of scale and orientation.}
        \label{fig:results}
	\end{center}%
}]

\footnotetext{$^\ast$ Work done while at Amazon}
\footnotetext{$^\dagger$ Equal contributions}

\begin{abstract}
    \vspace{-1em}
This work focuses on low bitrate video streaming scenarios (e.g. $50$ -- $200$Kbps) where the video quality is severely compromised.
We present a family of novel deep generative models for enhancing perceptual video quality of such streams by performing super-resolution while also removing compression artifacts. 
Our model, which we call SuperTran, consumes as input a single high-quality, high-resolution reference images in addition to the low-quality, low-resolution video stream. 
The model thus learns how to borrow or copy visual elements like textures from the reference image and fill in the remaining details from the low resolution stream in order to produce perceptually enhanced output video. 
The reference frame can be sent once at the start of the video session or be retrieved from a gallery.
Importantly, the resulting output has substantially better detail than what has been otherwise possible with methods that only use a low resolution input such as the SuperVEGAN method.
SuperTran works in real-time (upto 30 frames/sec) on the cloud alongside standard pipelines.
\end{abstract}
\vspace{-6mm}



\section{Introduction and Related Art}
\label{sec:relatedart}
Video transmission accounts for over $75\%$ of global IP traffic\cite{Cisco2019} and is only expected to increase with time.
This paper is addressing the problem of transmitting high resolution videos ($\geq$720p) under low bitrate (e.g. $\leq$200Kbps) settings. 
As the adoption of mobile phones having high-quality cameras continues to increase, such low-bandwidth, high-demand circumstances occur quite often, spanning geographies.

Video super-resolution and enhacement has been a long studied task with a rich history of work.
We only discuss representative works here and refer interested readers to \cite{anwar2019deep,rao2012survey, wang2019deep} for excellent surveys of techniques for super-resolution and overall enhancement of videos.

{\bf Video super resolution.}
Deep learning based techniques have dominated benchmarks on this task.
Recent work has shown that significant improvement can be made by accounting for motion implicitly\cite{Jo:etal:2018} or explicitly~\cite{Caballero:2017, Mehdi:etal:FRVSR:2018, supervegan}.
Generation of perceptually realistic details in the outputs\cite{Mengyu:etal:2018,Lucas:etal:2018,Pellitero:etal:2018,supervegan} has been with use of GANs\cite{Goodfellow2014GenerativeAN}.

{\bf Video Enhancement.} 
Works on enhancement have tried to tackle correction of compression related spatial and temporal artifacts such as blocking, ringing, flickering, and color changes from reduced color depth with decreasing bitrate.
Deblocking has been tackled using design of targetted correction filters\cite{Sung:etal:1999, List:etal:2003}, Kalman filters\cite{Lu_2018_ECCV}, and optical flow\cite{Bao:etal:2018,Xue:etal:2017}.
Some works performed video enhancement in conjunction with super-resolution by incorporating frame alignment modules\cite{Wang:etal:EDVR:2019}, offline detection of non-PQF's (Peak Quality Frames)\cite{Yang:etal:MFQE:2018} and combination of learned upsampling filters with GANs\cite{supervegan}.

{\bf Reference-based methods.}
Adjacent to the previous approaches, there is a body of work focused on harnessing additional reference imagery in order to improve performance on tasks like colorization and super-resolution.
Alignment based approaches~\cite{wang2016light, yue2013landmark, zheng2018crossnet, yue2013landmark} focus on aligning the reference and low-resolution images via a registration procedure\cite{yue2013landmark}, via warping based feature synthesis~\cite{wang2016light} or using optical flow~\cite{zheng2018crossnet}.

Some methods~\cite{boominathan2014improving, zhang2019image, zheng2017learning} adopt a patch-based alignment procedure to extract detail from reference images via local matching. 
In addition to the pixel space, this can also be done on image gradients~\cite{boominathan2014improving} and semantic features from CNNs~\cite{zheng2017learning, simonyan2014very, Yang_2020}.
These prior works are restricted to super-resolution and operate on single images, not videos.
SuperTran is aimed at improving upon video enhancement with aid from reference imagery.

\section{Methodology}
\vspace{-0.5em}
A strategy for dealing with low bitrates is to transmit videos at a reduced resolution and compensate for the quality by performing enhancement later\cite{supervegan}.
To the best of our knowledge, SuperTran is the first model to jointly tackle super resolution, video deblocking and overall video enhancement with a reference-based approach.

\begin{figure}
    \includegraphics[width=0.48\textwidth]{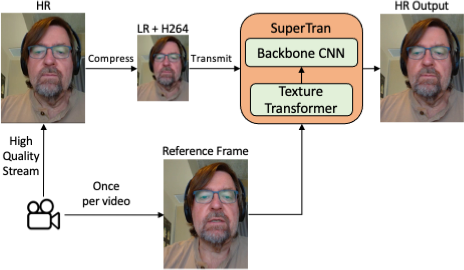}
    \vspace{-1.2em}
    \caption{\textbf{System figure.}
    Device can record high resolution (HR) video but only transmits at low resolution (LR) after compressing with H264 codec due to limited bandwidth. 
    Additionally, the device transmits a single high-resolution image (Ref) at the start of the video session.
    SuperTran consumes these as input and produces an enhanced, artifact-free video as output in real-time. 
    }  
    \vspace{-1.5em}
	\label{fig:system} 
\end{figure}

Given a sequence of $2k+1$ consecutive low-resolution, low-quality frames $I^{in}_{[t-k : t+k]}$ from a video stream, the goal of video enhancement is to estimate a high-resolution, high-quality frame $I^{out}$ corresponding to $I^{in}_{t}$.
SuperTran uses a reference frame $I^{ref}$ as additional input.
Our system to achieve this task is illustrated in Figure \ref{fig:system}.
SuperTran consists of two main modules, namely the Backbone network and the Multi-Scale Texture Transformer network (MSTT).

\vspace{-1.7em}
\begin{figure*}
    \centering
    \includegraphics[width=1\textwidth]{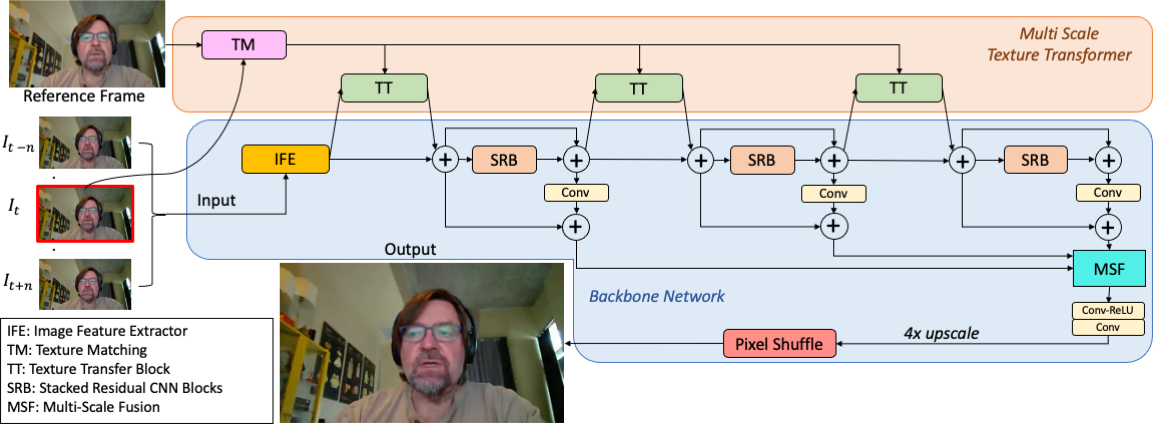}
    \vspace{-1.2em}
    \caption{\textbf{Model architecture.} The SuperTran model architecture consists of a main, feedforward CNN with multiple Texture Transformer modules interleaved at different stages of the model. The model received multiple frames as input capturing a temporal window and uses only the current frame $I_t$ for texture transfer from the reference image.} 
    \vspace{-1.5em}
    \label{fig:model} 
\end{figure*}

\vspace{0.3em}
\subsection{Backbone Network}
The backbone network is the central CNN architecture making up SuperTran. 
This network takes as input one or more low resolution, compressed input images (video frames) and produces as output an enhanced version of the central input frame. 
The backbone network receives as additional input the features processed by the Transformer network at various stages. 
Consuming neighboring frames as input allows the model to learn temporal patterns in data.

The design of SuperTran is optimized for high inference speed.
Thus, all operations are performed at the low input image resolution. 
As shown in Figure \ref{fig:model}, SuperTran consists of a series of stacked residual blocks with MSTT modules interleaved periodically.
Features at multiple points in the network are collected and processed jointly to obtain the final upscaled output with a pixel shuffle layer.

\subsection{Multi-Scale Texture Transformer}
\vspace{-0.5em}
The goal of the texture transformer module is to extract textures from the reference image such that they closely match with the textures in the input image. 
Since the reference frame is of higher visual quality, the extracted features, when transferred, can aid the backbone network in producing improved output.
The texture transformer module consumes $I^{in}_t$, reference image $I^{ref}$ and its downscaled version $I^{ref\downarrow}$. 
For rescaling images, we use bicubic interpolation.
Note that while the Backbone network receives a set of input frames as input, the Transformer network only receives the current frame in addition to the reference frame.

\textbf{Learnable Texture Extractor.} The Learnable Texture Extractor (LTE) is a CNN that consumes an image and produces high dimensional feature representations for it at multiple scales. 
These features can be thought of as textures.
We utilize weight sharing for LTE and learn a joint feature space for $I^{in}_t$, $I^{ref}$ and $I^{ref\downarrow}$.
The outputs from LTE are treated as the query $Q$, key $K$, and value $V$ for the next step and are the key ingredients of the transformer module.

\textbf{Feature matching.} We perform patch-level similarity matching between $Q$ and $K$ by first extracting $d \times d$ size patches from both. 
The similarity (relevance) score $R_{i, j}$ is computed using a normalized inner product between patches $Q_i$ and $K_j$.
\begin{align}
    R_{i,j} = \left< \frac{Q_{i}}{\left\| Q_{i} \right\|}, \frac{K_{j}}{\left\| K_{j} \right\|} \right>
\end{align}

\textbf{Attention Mechanism.} Using the relevance scores, we transfer textures from reference image in two ways. 
We utilize hard attention to transfer only the most relevant (best match) textures and soft attention to transfer a synthesized blend of relevant textures.
For hard attention, we extract the indices $H_i$ for the most relevant matches and use them to obtain the transferred features $T$ by performing a selection operation on the patches $V$.
\begin{align}
    H_i &= \mathop{\arg\max}_{j} R_{i,j} & T_i = V_{H_i} & \forall i
\end{align}
The soft attention map $S$ is computed from the relevance scores to indicate confidence for the best matching texture.
\begin{align}
    S_i &= \mathop{\max}_{j} R_{i,j} 
\end{align}
As a final step, we concatenate the features from $I^{in}$ and the transferred features and apply elementwise soft attention to obtain blended features. 
The texture matching (TM) components described above along with texture transfer (TT) at multiple scales are shown in Figure \ref{fig:model}.

\subsection{Loss Function}
\vspace{-0.5em}
    We train SuperTran in a multi-objective optimization setting with a loss function consisting of 4 components: pixel-wise reconstruction loss $\mathcal{L}_{rec}$, adversarial loss $\mathcal{L}_{adv}$ , perceptual loss $\mathcal{L}_{per}$, and texture loss $\mathcal{L}_{tex}$. 
    For the definitions below, we consider $N = C * H * W$ where $C, H, W$ are the number of channels, height and width respectively.
    \begin{align}
        \mathcal{L} =  \lambda_{rec} \mathcal{L}_{rec} + \lambda_{adv} \mathcal{L}_{adv} + \lambda_{per} \mathcal{L}_{per} + 
        \lambda_{tex} \mathcal{L}_{tex}
    \end{align}
    \textbf{Reconstruction Loss.} This is measured with the per-pixel $L_1$ distance.
    \begin{align}
        \mathcal{L}_{rec}(I^{gt}, I^{out}) = \frac{1}{N} \left \| I^{gt} - I^{out} \right \|_1
    \end{align}
    \textbf{Adversarial Loss.} We use WGAN-GP\cite{gulrajani2017improved} as our GAN loss which uses a gradient norm penalty in addition to the Wasserstein GAN formulation\cite{arjovsky2017wasserstein}.
    The loss is given by,  
    \begin{align}
        \mathcal{L}_{D} &= \mathop{\mathbb{E}}\limits_{\tilde{x} \sim \mathbb{P}_g} \big[ D(\tilde{x}) \big] - \mathop{\mathbb{E}}\limits_{x \sim \mathbb{P}_r} \big[ D(x) \big] + \notag \\
        &\ \ \ \ \  \lambda \mathop{\mathbb{E}}\limits_{\hat{x} \sim \mathbb{P}_{\hat{x}}} \big[(\left \| \nabla_{\hat{x}} D(\hat{x}) \right \|_2 - 1)^2 \big], \\
        \mathcal{L}_{adv} &= - \mathop{\mathbb{E}}\limits_{\tilde{x} \sim \mathbb{P}_g}\big[ D(\tilde{x}) \big],
    \end{align}
    where $D$ is our discriminator, $x$, $\tilde{x}$ are samples from real and generated distributions respectively and $\mathbb{P}$ denotes the corresponding probability distribution.
    
    \textbf{Perceptual Loss.} Our perceptual loss consists of two terms measuring similarity in the feature space. In addition to the typical loss in the VGG19 feature space for output and ground truth image, we use an additional loss between the LTE transferred features and those of the output image. This ensures that $I^{out}$ retains the transferred features $T$.
    \begin{align}
        \mathcal{L}_{per} =& \frac{1}{N} \left \| \phi^{vgg}(I^{out}) - \phi^{vgg}(I^{gt})\right \|^2_2 + \notag \\
                          & \frac{1}{N} \left \| \phi^{lte}(I^{out}) - T  \right\|^2_2
    \end{align}
    Here, $\phi$ represents features extracted from intermediate network layers for VGG and LTE networks.

    \textbf{Texture Loss.} Similar to the perceptual loss, we compute this loss in the VGG19 space, but with a model that was trained for the task of texture classification on the MINC dataset. We observe that this loss brings stability to outputs.
    \begin{align}
        \mathcal{L}_{tex} =& \frac{1}{N} \left \| \phi^{minc}(I^{out}) - \phi^{minc}(I^{gt})\right \|^2_2 
    \end{align}
    Here, $\phi$ represents features extracted from intermediate network layers of the MINC trained VGG network.

\section{Experiments}
\textbf{Datasets.} 
For training, we collected 10 video sequences spanning 17 minutes each, recorded at 1080p resolution (i.e. $1920 \times 1080$px).
These videos depict outdoor scenery and are recorded from a device different from the test videos.
From these videos, we sample 20,000 frames and resize them to $1280 \times 720$px to make up the ground truth for training dataset. 
For testing, we have curated a Talking Heads dataset containing a collection of high quality videos of people speaking into the camera (on laptop or a handheld mobile device) in a casual format akin to a video calling setup.
These videos have been recorded under a diverse range of settings including varying lightning conditions, camera motion, surroundings, camera orientation, video resolution and the subjects (talking heads) in the video. 
The dataset consists of a total of 71 videos spanning over 1hour 36mins with an average video length of 1min 11sec.

To create paired data, we first bilineary downscale the videos to $320 \times 180$px and encode them using the H264 implementation in Gstreamer\cite{gstreamer}, modifying only the \textit{bitrate} parameter of the  encoder.
The downscaling factor (here, 4x) can be easily modified to derive multiple SuperTran variants.
Note that our framework is not restricted to the H264 video codec and can adapt to other codecs with corresponding training.

\textbf{Training Setup.}
We train using randomly sampled $96 \times 96$px patches from the low-resolution, compressed image dataset and use a window of 5 frames as input.
The model is tasked to produce a 4x upscaled output corresponding to the middle input frame.
For the loss function, we set $\lambda_{rec}$, $\lambda_{adv}$, $\lambda_{per}$, $\lambda_{tex}$ to $1$, $5e^{-4}$, $1e^{-2}$ and $1e^{-2}$ respectively.
We initialize the learning rate to $1e^{-3}$ and train the model for $500$ epochs halving the learning rate after $300$ epochs.
For the first $20$ epochs, we train using only reconstruction loss and enable other losses thereafter. 

\textbf{Metrics.} 
We adopt metrics used in other works: PSNR, SSIM\cite{wang2004image} and LPIPS\cite{zhang2018perceptual}. 
PSNR and SSIM are the most common metrics used for assessing image quality, but they do not fully capture the perceptual quality of the image and favor blurriness. 
We mainly target the LPIPS metric, which is a learned metric in the CNN feature space and is optimized for high correlation with human judgement of perceptual details in images.
The LPIPS metric was also found to correlate well for human evaluation of videos\cite{supervegan}. 

\textbf{Results.}
We report quantitative results averaged over all videos in our test dataset in Table \ref{tab:results}.
For comparison, we consider the H264 codec with bicubic upscaling as the baseline and SuperVEGAN\cite{supervegan} as the representative state-of-the-art technique.
We perform evaluation using inputs of resolution $320 \times 180$px and compute metrics at the 4x upscaled output resolution $1280 \times 720$px.
For illustrative purposes, we show qualitative results on outdoor scenes in addition to our test data in Figure \ref{fig:results}.
Please refer the supplementary material for video results.
\vspace{-0.5em}
\begin{table}[h!]
    \footnotesize
    \setlength{\tabcolsep}{5pt}
    \begin{center}
    \begin{tabular}{|c|c|c|c|c|}
    \hline 
    Method & Bitrate & PSNR$\uparrow$ & LPIPS$\downarrow$ & SSIM$\uparrow$\\
    \hline \hline
    H264 & 50 & $27.43 \pm 3.56$ & $0.33 \pm 0.11$ &  $0.82 \pm 0.08$ \\
    SVEGAN\cite{supervegan} & 50 & $27.25 \pm 3.37$ & $0.23 \pm 0.11$ &  $0.81 \pm 0.08$ \\
    SuperTran & 50 & $26.32 \pm 2.82$ & $0.18 \pm 0.11$ & $0.82 \pm 0.08$ \\
    \hline
    H264 & 100 & $29.00 \pm 2.66$ & $0.29 \pm 0.08$ & $0.84 \pm 0.06$ \\
    SVEGAN\cite{supervegan} & 100 & $28.71 \pm 2.42$ & $0.18 \pm 0.07$ & $0.84 \pm 0.06$ \\
    SuperTran & 100 & $27.6 \pm 1.94$ & $0.13 \pm 0.06$ & $0.84 \pm 0.06$ \\
    \hline
    H264 & 200 & $30.03 \pm 2.48$ & $0.26 \pm 0.07$ & $0.86 \pm 0.02$ \\
    SVEGAN\cite{supervegan} & 200 & $29.52 \pm 2.17$ & $0.15 \pm 0.05$ & $0.85 \pm 0.05$ \\
    SuperTran & 200 & $28.14 \pm 1.78$ & $0.11 \pm 0.04$ & $0.86 \pm 0.05$ \\
    \hline
    \end{tabular}
    \end{center}
    \vspace{-1.5em}
    \caption{Results at 50, 100 and 200kbps bitrate. 
    For PSNR and SSIM, higher is better. 
    For LPIPS, lower is better.
    SuperTran obtains up to $27\%$ relative improvement in LPIPS over SVEGAN.}
    \label{tab:results}
    \vspace{-1em}
\end{table}
\vspace{-1.25em}

\section{Conclusion}
\vspace{-0.5em}
We introduce a novel video transformer for enhancing low bitrate video
streams with the aid of reference imagery. Our proposed
model architecture, SuperTran, is able to borrow visual detail
from a single reference image and remove artifacts from video
streams significantly better than with previous methods. SuperTran
is domain agnostic in that it is not restricted to a specific
application (e.g., human faces) unlike recent work\cite{maxine}. Our cloud-based implementation of SuperTran
works on live video at up to 30fps at 720p resolution and can support a variety
of video streaming, processing and storage applications.

{\small
\bibliographystyle{ieee_fullname}
\bibliography{references}
}

\end{document}